# Clustering Students Based on Gamification User Types and Learning Styles

**Emre Arslan**
emrearslan1308@gmail.com
Marmara University
Istanbul, Türkiye

**Atilla Özkaymak**
atilla-serkan@hotmail.com
Marmara University
Istanbul, Türkiye

**Nesrin Özdener Dönmez**
nozdener@marmara.edu.tr
Marmara University
Istanbul, Türkiye

**Abstract**

The aim of this study is clustering students according to their gamification user types and learning styles with the purpose of providing instructors with a new perspective of grouping students in case of clustering which cannot be done by hand when there are multiple scales in data. The data used consists of 251 students who were enrolled at a Turkish state university. When grouping students, K-means algorithm has been utilized as clustering algorithm. As for determining the gamification user types and learning styles of students, Gamification User Type Hexad Scale and Grasha-Riechmann Student Learning Style Scale have been used respectively. Silhouette coefficient is utilized as clustering quality measure. After fitting the algorithm in several ways, highest Silhouette coefficient obtained was 0.12 meaning that results are neutral but not satisfactory. All the statistical operations and data visualizations were made using Python programming language.

## 1. Introduction

People have started to realize the importance of e-learning systems as they were compelled to attend or perform online classes after the breakout of COVID-19 pandemic. Under these circumstances, adaptive technologies have been wielded by many institutes with the objective of creating appropriate education environments where students can experience better learning gains [1]. Adaptive educational systems enhance and support the learning process monitoring learners' characteristics and making adjustments accordingly [2].

As Shute and Zapata-Rivera stated [3], utilization of adaptive technologies requires identifying students' characteristics—e.g. level of knowledge, skills, personality traits—as accurately as possible. Regarding this identification phase, some students may have more than one character or learning trait when measured by relevant scales. Also, there is another potential problem which comes in when there are more than one scale applied to students in order to group them. This grouping process may possibly be done by hand if the quantity of students is relatively small and there is only one scale applied however it may be hard enough to group them as the number of students is getting larger and there are multiple scales. To address these problems, machine learning techniques can be used for facilitating the grouping and reducing the workload that instructors undertake.

The vast chunk of the industry is commencing to utilize the machine learning algorithms to enhance the quality of their service. In the past, researchers have done studies [4, 5] on the usage of clustering from which it can be also inferred that many research papers are getting published about implementation of machine learning algorithms into education like Merceron & Yacef [6] did on clustering students to help instructors in the evaluation phase. In this context, this study can provide a new perspective of grouping students for instructors.

### 1.1 Limitations

- The data used in this research is limited to students in Türkiye.
- Due to the randomness in the nature of K-means algorithm, the clusters obtained may not always be consistent. Meaning that, students in the same cluster may not be placed into the same cluster if the K-means algorithm is run again.

### 1.2 Working Mechanism of K-means Algorithm

The algorithm that has been chosen for this study is K-means algorithm as it is commonly used for clustering tasks because of its being more general-purpose. The logic which K-means is built upon is initializing K number of data points—also known as centroids—and sliding those points until finding the optimal place. There are two main steps that K-means algorithm takes; initialization and adjustment. As initialization, we have used k-means++ which is a smart centroid initialization method proposed by Arthur and Vassilvitskii [7] for getting more consistent clusters. The algorithm proposed is as follows where $d(x)$ is Euclidean distance, $x$ is observation and $X$ represents all data points:

1. Choose one of the observations randomly as first centroid.
2. Compute the $d(x)^2$ between each observation and the centroid.
3. Add a new centroid to the place of observation where the probability calculated with

$$\frac{d(x)^2}{\sum_{x \in X} d(x)^2} \quad (1)$$

   is maximum.
4. Repeat the step 2 and step 3 by calculating the $d(x)^2$ between each observation and nearest cluster until reaching the specified number of clusters—K value.

After initialization part, outline of the way K-means algorithm works—adjustment section—is as follows:

1. Compute the $d(x)^2$ between each observation and centroids.
2. Assign each observation to the nearest centroid.
3. Calculate the mean of each cluster.
4. Slide the centroids to the clusters' mean point.
5. Repeat previous steps until centroids are no longer getting slided.

## 2. Related Work

With significant improvements in artificial intelligence technology, the use of unsupervised learning methods has become popular among instructional technologists as it also serves to educational purposes. This being the case, researchers have conducted studies using unsupervised learning techniques such as clustering students for helping teachers in the evaluation part [6], clustering moodle data for profiling students [8], clustering lifestyle factors—physical activity, alcohol consumption, diet quality and smoking habit—of students [9] and so on. But the ones that are closely related to this research paper are those in one of which researchers tried to cluster students according to their learning style [4] and in another students have been clustered based on their gamification behavior [5].

However, as far as we could see, in most of the research papers on clustering students researchers have used the data in which the scales that provide information about the educational tendencies of students are found but the data which clustering was done according to consisted of one element. Meaning that, students got clustered according to the criteria which give information regarding one characteristic side of students—e.g. gamification behavior, learning styles or pior knowledge [10]. Also, there have been some works [11, 12] that clustered students based on multiple criteria that indicate students preferences, knowledge or learning characteristics but they are not so many. In this context, this paper may contribute to the literature of this kind of clustering. Also, this paper is written in such a way that researchers who want to utilize unsupervised learning techniques in their research can clearly see the workflow of clustering process.

## 3. Method

The type of this research is applied research which aims at clustering students according to their gamification user type and learning style by making use of unsupervised learning methods.

### 3.1 Data Collection

The data used in this study comprises two datasets containing each student's both gamification user type and learning style. Gamification User Type Hexad Scale has been used as scale for determining the students' gamification type [13] which consisted of 24 questions on 7-point Likert scale. Because the data was collected in Turkey, Turkish version [14] of this scale was applied to students. For obtaining the learning styles of students, Grasha-Riechmann Student Learning Style Scale (GRSLSS) [15] has been used as learning style scale that is comprised of 60 questions on 5-point Likert scale and this part of the dataset is collected by applying the adaptation of GRSLSS into Turkish language [16].

During the whole process, convenience method was used as data collection method. Relevant scales were applied to university students via an online questionnaire as Google Form document. Finally, each student's gamification user type—Philanthropist, Socializer, Free Spirit, Achiever, Disruptor or Player—and learning style—Independent, Avoidant, Collaborative, Dependent,

Competitive or Participant—have been calculated according to the instructions that scale creators have given.

**3.2 Participants**

Participants are consisting of 251 students at a Turkish state university, 166 of whom were female and 85 of whom were male. The participants have been determined using the convenience sampling method.

**3.3 Data Mining**

Prior to clustering, the data should be preprocessed so that the data is in the form that clustering algorithm can work with. Before moving on to the data mining part, it would be more appropriate to introduce the readers of this paper to the workflow that has been followed during the clustering process. The workflow includes data standardization, dimensionality reduction, clustering and analysis of clusters in order.

**3.3.1 Data Standardization**

Because the scales applied are on different ranges, Gamification User Type Hexad Scale [13] is on 7-point Likert scale and Grasha-Reichmann Learning Style Scale [15] is on 5-point Likert scale, there is a need of data standardization—putting all the features of data on the same scale. Another reason for standardizing the data is that because K-means algorithm uses Euclidean distance as distance measure, data should be standardized for features containing large numbers not to overpower the features having small numbers [17]. There is a wide range of normalization methods such as Min-Max normalization, Z-score standardization and Decimal Scaling normalization. In this section, we have standardized our data using Z-score formula because of the results showing that Z-score Standardization is more efficient and effective than Min-Max normalization and Decimal Scaling normalization when clustering with K-means algorithm [17]. Z-score standardization sets the mean to 0 and standard deviation to 1 so that data features are on the same scale and in the shape of Gaussian distribution, i.e. normal distribution. The formula used for standardizing the data is as follows:

$$z = \frac{x - \mu}{\sigma} \quad (2)$$

**3.3.2 Dimensionality Reduction**

The data we used in this research paper consists of 84 variables— or dimensions. As this being the case, there is a problematic situation that should be handled: curse of dimensionality. Curse of dimensionality is basically occuring where there is large number of variables and it affects the behavior and performance of learning algorithms negatively [18]. As to handle this situation, we have utilized Principal Component Analysis which is a widely used technique of dimensionality reduction. According to the Hair [19], in social sciences it is not uncommon to consider a solution that accounts for 60% of the total variance as satisfactory. Considering the opinion of Hair, we have decided to use first 25 principal components that explains over 70% of total variance present in the data. In the Figure 1, the variance that each component accounts for and cumulative sum of explained variance can be seen:

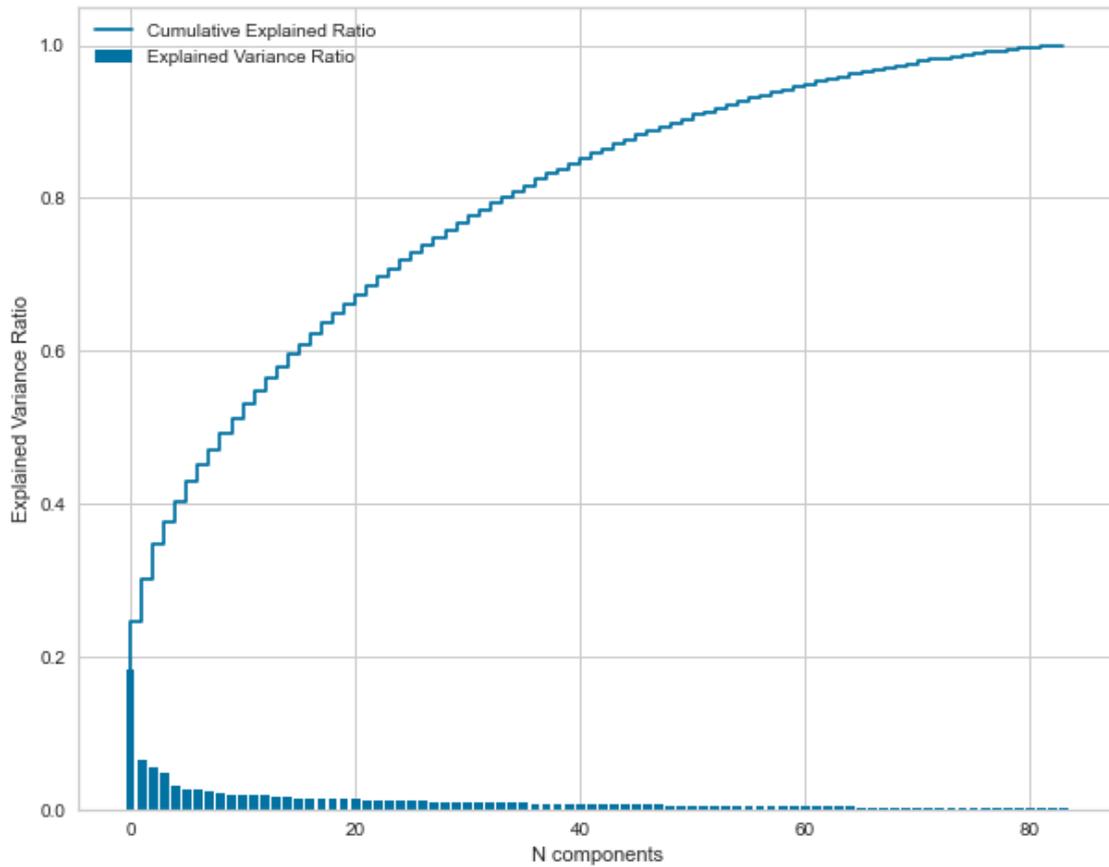

**Figure 1.** Variance that components account for and cumulative sum of explained variance

### 3.3.3 Clustering

As it has been explained in detail earlier in this paper, K-means algorithm basically partitions the data into groups of patterns but the tricky part is selecting the K value which is the number of groups that students will be separated into. To overcome this part, we have used Elbow method which shows the best number of clusters to choose by calculating the Sum of Squared Error in each cluster between observations and their closest centroid [20]. As the number of K increases, the variation within each cluster and hence the mean of Sum of Squared Error is decreasing. This decrease in the mean of Sum of Squared Error, or distortion, for each K value is very fast in the very beginning but after some point it starts to slow down. Where this is happening depicts an elbow and it shows the number of K that is appropriate to choose. For the sake of convenience, we have utilized the KElbowVisualizer tool of Yellowbrick library to easily see the elbow and choosing K as 5 is suggested by the tool. In the Figure 2, the elbow can be seen:

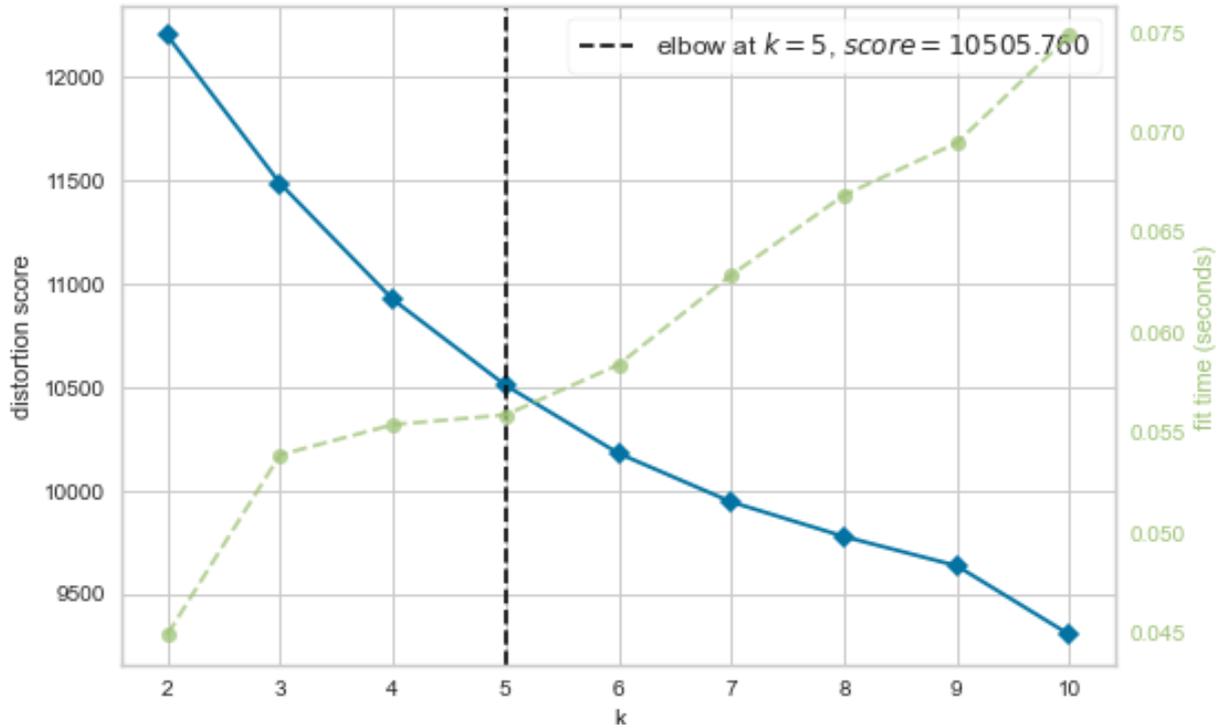

**Figure 2.** Distortion score for each k value and suggested elbow point

### 3.3.4 Analysis of Clusters

Once the clustering process is done, it is time to analyze each cluster. As it has been expressed, K value was chosen as 5 which means the students were separated into 5 clusters. Firstly, we have taken a look at the distribution of students and it is observed that Cluster 4 has the largest number of students with 69 students while Cluster 1 having the smallest number with 43 students. The distribution of the students can be seen in the Table 1 and Figure 3:

**Table 1.** Distribution of students by cluster

| Cluster | Number of Students |
| --- | --- |
| Cluster 4 | 69 |
| Cluster 5 | 47 |
| Cluster 2 | 46 |
| Cluster 3 | 46 |
| Cluster 1 | 43 |

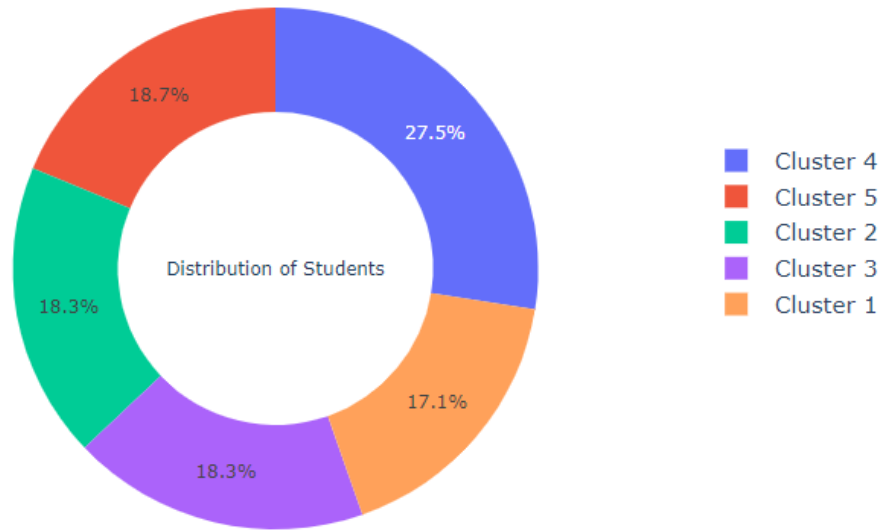

**Figure 3.** Visualization of distribution of students by cluster

The means of all the students' gamification user type and learning style scores have been shown in the Table 2 which contains the descriptive statistics. It can be seen that students show achiever characteristics while having nearly same scores in two learning styles—independent and dependent—if we take the integer part into account. In Table 2, gamification user type is abbreviated as GUT and learning style as LS:

**Table 2.** Descriptive statistics of learning style and gamification user type scores of all students

| LS/GUT | Count | Mean | Standard Deviation |
| --- | --- | --- | --- |
| Independent (LS) | 251 | 40.11 | 5.09 |
| Avoidant (LS) | 251 | 26.97 | 5.56 |
| Collaborative (LS) | 251 | 38.72 | 6.69 |
| Dependent (LS) | 251 | 40.55 | 4.29 |
| Competitive (LS) | 251 | 32.01 | 7.70 |
| Participant (LS) | 251 | 35.63 | 6.42 |
| Philanthropist (GUT) | 251 | 24.29 | 3.61 |
| Socializer (GUT) | 251 | 21.51 | 4.86 |
| Free Spirit (GUT) | 251 | 24.85 | 2.73 |
| Achiever (GUT) | 251 | 25.34 | 2.90 |
| Disruptor (GUT) | 251 | 12.44 | 5.14 |
| Player (GUT) | 251 | 23.48 | 4.05 |

Each cluster's mean scores of gamification user type and learning style is shown in the Table 3. Taking the integer part into the account, it can be observed that:

- **Cluster 1** has the Independent learning style and is Achiever as gamification user type
- **Cluster 4** has the Collaborative learning style as it has two gamification user types; Philanthropist and Achiever.
- **Cluster 2** has the Collaborative learning style while having Achiever gamification user type
- **Cluster 3** has the Dependent learning style as it shows Achiever traits as gamification user type
- **Cluster 5** carries the Independent characteristics as learning style while it has three gamification user types; Free Spirit, Achiever and Player.

**Table 3.** Learning style and gamification user type scores of all students.

| LS/GUT | Cluster-1 | Cluster-2 | Cluster-3 | Cluster-4 | Cluster-5 |
|---|---|---|---|---|---|
| Independent (LS) | 41.83 | 43.41 | 34.50 | 38.44 | 43.27 |
| Avoidant (LS) | 33.34 | 23.39 | 30.06 | 23.81 | 26.27 |
| Collaborative (LS) | 35.93 | 46.04 | 31.30 | 41.78 | 36.87 |
| Dependent (LS) | 38.95 | 43.91 | 36.02 | 40.76 | 42.85 |
| Competitive (LS) | 26.72 | 39.58 | 26.65 | 30.01 | 37.65 |
| Participant (LS) | 29.11 | 42.56 | 29.84 | 36.82 | 38.74 |
| Philanthropist (GUT) | 24.25 | 26.69 | 20.89 | 25.52 | 23.53 |
| Socializer (GUT) | 21.20 | 25.28 | 18.08 | 22.75 | 19.63 |
| Free Spirit (GUT) | 25.69 | 26.39 | 21.82 | 24.89 | 25.48 |
| Achiever (GUT) | 26.09 | 27.45 | 22.15 | 25.30 | 25.76 |
| Disruptor (GUT) | 16.44 | 11.41 | 12.32 | 9.86 | 13.68 |
| Player (GUT) | 24.32 | 25.82 | 20.04 | 22.02 | 25.91 |

To measure the quality of clustering, we have used silhouette coefficient which provides information about how similar the data is to its assigned cluster compared to other clusters [21]. Silhouette coefficient is calculated using the following formula where $a$ represents the average intra-cluster distance—distance between a data point and other data points belonging to same cluster—and $b$ represents the average distance between a data point and other points in nearest cluster.

$$s = \frac{b - a}{\max(a, b)} \quad (3)$$

The silhouette coefficient ranges in $[-1, 1]$ where -1 denotes that clustering is done wrongly whereas 1 denotes good clustering, well separated clusters, as the score that we obtained is 0.04. Since we have fit K-means algorithm with the student answers, not with calculated learning style or gamification user type scores, we have also tried to fit the algorithm with learning style and gamification user type scores without giving the answers of students—PCA was also applied taking the first 5 components which account for over 75% of the total variance—and the silhouette coefficient yielded was 0.12.

Because students' answers' ranges are $[1, 7]$ and $[1, 5]$ in variables in which there is an order between these values, the data may be regarded as ordinal categorical data. As Riani, Torti and Zani [22] stated, it is very difficult to detect outliers in an ordinal categorical variable when the term outlier is defined as the observations different from the majority of observations since a value can't be considered as an outlier because of the fact that observations can have values from 1 to n. However, in special cases there can possibly be the presence of outliers regarding the frequency distribution of a variable. Just in case, we have clustered our data using robust scaler method which scales the data in such a way that features become more robust to outliers removing the median and scaling the data according to the Interquartile Range. With application of this scaling in data standardization part, the silhouette coefficient obtained was 0.03.

## 4. Discussion and Conclusion

This study was conducted for providing instructors with a new perspective of grouping students when there is a need of clustering students which cannot be done by hand because of the dataset size and number of scales that students get clustered according to. In the phase of determining the clustering quality, silhouette coefficient that we got is 0.04 which is positive, but this score is quite close to 0. Meaning that, clusters have a high tendency to overlap. This problem may be caused by having more than one scales to cluster according to as relatively better silhouette coefficients were obtained in related studies where there was one criterion used—student activity in one and student behavior in another—to cluster students based on [23, 24].

When we have fit the algorithm with learning style and gamification user type scores instead of fitting with answers of the students, the silhouette coefficient was 0.12 which is relatively better than what we got when the algorithm was fit with students' answers—silhouette coefficient of 0.04—however this score is still substantially close to 0 indicating that clusters' overlapping problem continues. After trying to scale the data using Robust Scaler in case of outlier presence concerning the frequency distribution of variables, the quality of clustering has even deteriorated.

In this paper, we have tried to cluster students according to their gamification user types and learning styles with the objective of demonstrating that unsupervised learning techniques can be an asset for instructors when clustering students according to more than one scale describing students' learning styles and behavior in educational games. The results regarding the quality of clustering are neutral and not satisfactory. The reasons could be these: the scales applied to the students may not be appropriate enough for this kind of clustering, K-means algorithm may not be doing well on such data where the data consists of ordinal categorical variables, data preprocessing may not be done suitably. Future work on such clustering should be done using different multiple criteria that provide information about students' educational characteristics utilizing other unsupervised learning algorithms.